\documentclass[letterpaper]{article} 
\usepackage{aaai23}  
\usepackage{times}  
\usepackage{helvet}  
\usepackage{courier}  
\usepackage[hyphens]{url}  
\usepackage{graphicx} 
\def\UrlFont{\rm}  
\usepackage{natbib}  
\usepackage{caption} 
\usepackage[dvipsnames]{xcolor}
\usepackage{colortbl}
\usepackage{zref-totpages}
\usepackage{tabularx,colortbl}
\usepackage{algorithm}
\usepackage{algorithmic}

\usepackage{newfloat}
\usepackage{listings}
\DeclareCaptionStyle{ruled}{labelfont=normalfont,labelsep=colon,strut=off} 
\floatstyle{ruled}
\newfloat{listing}{tb}{lst}{}
\floatname{listing}{Listing}
\usepackage{booktabs}
\usepackage{subfiles}

\title{Not Enough Labeled Data? Just Add Semantics: A Data-Efficient Method for Inferring Online Health Texts}
\author{Joseph Gatto, Sarah M. Preum}
\affiliations{
    Department of Computer Science, Dartmouth College
    
}

\begin{document}
    
\maketitle

\begin{abstract}
User-generated texts available on the web and social platforms are often long and semantically challenging, making them difficult to annotate.  Obtaining human annotation becomes increasingly difficult as problem domains become more specialized. For example, many health NLP problems require domain experts to be a part of the annotation pipeline. Thus, it is crucial that we develop \textbf{low-resource NLP solutions} able to work with this set of limited-data problems. 

In this study, we employ Abstract Meaning Representation (AMR) graphs as a means to model low-resource Health NLP tasks sourced from various online health resources and communities. AMRs are well suited to model online health texts as they can represent multi-sentence inputs, abstract away from complex terminology, and model long-distance relationships between co-referring tokens. AMRs thus improve the ability of pre-trained language models to reason about high-complexity texts. Our experiments show that we can improve performance on 6 low-resource health NLP tasks by augmenting text embeddings with semantic graph embeddings. Our approach is task agnostic and easy to merge into any standard text classification pipeline.  We experimentally validate that AMRs are useful in the modeling of complex texts by analyzing performance through the lens of two textual complexity measures: the Flesch Kincaid Reading Level and Syntactic Complexity. Our error analysis shows that AMR-infused language models perform better on complex texts and generally show less predictive variance in the presence of changing complexity.

\end{abstract}

\section{Introduction}
In recent years, fine-tuning pre-trained language models (PLMs) has become a standard approach to text classification \cite{devlin2018bert,liu2019roberta}. However, it is becoming clear that many tasks are too complex to be modeled using the standard fine-tuning pipeline and require a more intricate, task-specific solution. This is particularly true in the domain of low-resource NLP --- the set of tasks where large-scale human-annotated data is unavailable. An exact definition of what makes a problem low-resource varies throughout the literature \cite{low_resource_survey}, and is often both problem and domain-specific. However, it is clear that when a certain language is underrepresented \cite{low_r_survey_2}, or a specific type of problem is very difficult to obtain annotation for \cite{10.1162/coli_a_00425}, a low-resource solution can improve performance on a given problem.

Limited annotation is common amongst datasets grounded in web and social media data \cite{kawintiranon2021knowledge, suicide_risk_reddit, Khanpour_Caragea_Biyani_2018}. This is due to the complex nature of the data as they are often heterogeneous, multi-sentence texts, making them challenging to annotate. Additional degrees of complexity may be added  for specific domains. For example, web and social texts regarding \textit{health} bring about more significant annotation challenges, as domain expertise is required for many health annotation tasks. Works supporting low-resource health NLP tasks have become increasingly common in recent years. However, many of these works focus on clinical health texts \cite{alsentzer-etal-2019-publicly, rohanian2022using}. Related works also lie in the field of health data augmentation \cite{ansari-etal-2021-data, yang-etal-2023-data}. However, task-agnostic augmentation methods have been shown to have limited scope for boosting PLM performance \cite{task-agnostic-data-augmentation} with specific examples of augmentation challenges for complex online health texts being discussed in  \cite{gatto2023HCD}. 

Limited work has been done about generalizable solutions to modeling \textit{online health texts}, i.e., healthcare texts specific to online health resources, communities, and social media, which we argue require explicit NLP solutions as they contain their unique linguistic traits. For example, the context within which various health terms are discussed on public platforms constantly evolves, making online textual health data semantically challenging to model. Additionally, most transformers are pre-trained with formal or grammatically correct language and are not inherently well-suited to user-generated texts. Finally, web and social health platforms encourage texts which are often multi-sentence, which pose unique NLP modeling challenges for understanding long-range dependency information between co-referring tokens. To address these challenges, in this study, we propose using Abstract Meaning Representation (AMR) graphs to model complex web and social health texts.

\begin{figure}[t]
    \centering
    \includegraphics[width=\columnwidth]{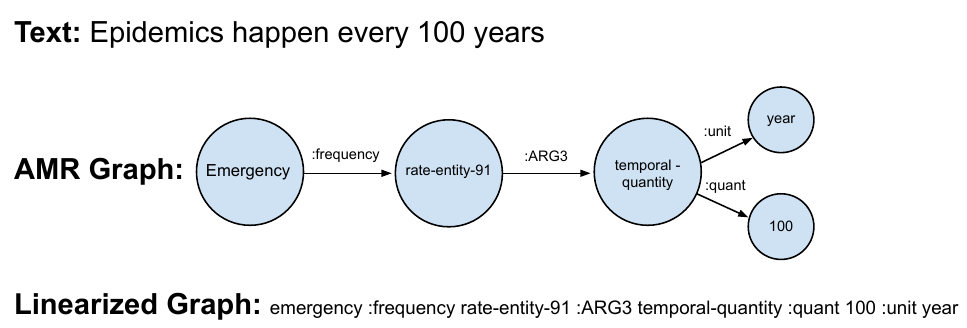}
    \caption{Example AMR graph and corresponding linearization for a given text. The AMR graph abstracts away from the text into high-level semantic concepts connected by semantic relations. The linearized version of the graph is a depth-first traversal of the AMR that allows the AMR to be read by standard transformer models. }
    \label{fig:example_parse}
\end{figure}

Abstract Meaning Representation (AMR) graphs have become a popular  semantic structure used in a variety of NLP tasks  \cite{bonial2020dialogue, amr_da, li-etal-2022-cross-domain}. Unlike other linguistic modeling tools such as Semantic Role Labeling (SRL) or Dependency Trees (DTs), AMRs abstract \textit{away} from the text, representing meaning using only high-level semantic concepts from a fixed vocabulary. Figure \ref{fig:example_parse} provides an example AMR graph. The graph contains mostly abstract concepts instead of actual words from the original sentence.

In this study, we show how we can leverage AMR graphs to improve performance on various low-resource health classification tasks by augmenting text embeddings with AMR information. We define a low-resource health task as one with $< 6,000$ human-annotated samples. This is a conservative definition of low-resource for health texts consistent with related literature in NLP and web mining \cite{ansari-etal-2021-data, MQP, covid_rumor, gatto2023HCD}. AMRs are an intuitive choice for the modeling of low-resource online health texts as AMRs provide a compact representation of the input space while explicitly modeling co-reference of multi-sentence texts \cite{ogorman-etal-2018-amr}. This makes AMRs well-suited to model the nuances of health texts. Additionally, AMRs can abstract away from complex health terminology. Consider the example in Figure \ref{fig:example_parse}, where ``Epidemics" is abstracted away to ``Emergency". Such abstraction can make BERT-based models more generalizable for domain-specific tasks. AMRs are also easy to integrate into existing transformer-based NLP pipelines. Finally, AMRs can be adapted to other languages, making them more inclusive than other English-only semantic structures. 

We validate our claim that AMRs aid in the modeling of complex health texts by investigating how performance changes with respect to textual complexity. We do this through the lens of two different complexity metrics 1) Estimated Reading Level \cite{flesch1948new}, a statistical measure of textual reading difficulty and 2) Syntactic Complexity \cite{dep_distance_ref}, a measure of the average difference between co-referring tokens in a given text. Our results show that AMR-infused PLMs exhibit better performance on texts with higher degrees of textual complexity when compared to an off-the-shelf text-only model. \\

\noindent A summary of our contributions are as follows:

\begin{enumerate}
    \item We show that AMR-augmented classifiers can increase performance across various low-resource health NLP domains including medical advice modeling, telemedicine, and the modeling of medical research literature. Augmenting text classifiers with AMR embeddings is shown to, on average, provide a 3-pt increase in F1 score to frozen text encoders and 1-pt increase in F1 score to dynamic end-to-end text encoders across all tasks. 
    
    \item We provide a thorough error analysis of our solution through the lens of textual complexity. We verify our claim that AMRs improve the modeling of complex texts by analyzing model performance on samples at  varying reading levels and degrees of syntactic complexity. Our analysis concludes that AMR-based models outperform text-only models on high-complexity health texts. 
     
\end{enumerate}

\section{Background and Related Works}
\subsection{Low-Resource Health NLP}
In recent years, the machine learning community has seen an emergence of important Health NLP problems. Unfortunately, many tasks in this space are difficult to obtain large-scale annotation for and live in the domain of low-resource NLP. For example, the onset of the COVID-19 pandemic inspired many researchers to construct COVID-specific misinformation datasets from web articles and social media posts  \cite{FA_COV}. However, given the difficulty of annotating health texts, it often appears that the best-case scenario for many of these problems is to obtain a few thousand ground truth, human-annotated samples, with many containing less than 1,000 annotations \cite{CoAID, monant}. Such little annotation hinders the use of various deep NLP architectures to solve these problems. Many other examples of low-resource health datasets can be found in the domain of public health entity extraction \cite{HealthE}, predicting suicide risk \cite{suicide_risk_reddit}, and detecting adverse drug reactions on Twitter \cite{TwitterADR}. In this work, we aim to study a diverse set of low-resource text classification tasks in the domain of Health NLP and how we can reduce the need for large numbers of human-annotated examples.

\subsection{Semantically Grounded Transformer Models}
The explicit modeling of semantics in transformer-based architectures has become a popular way to increase their modeling capacity. For example, in \cite{SemBERT} they show that augmenting BERT with Semantic Role Label information can improve performance on a variety of tasks within the GLUE benchmark. In \cite{srl_paraphrase}, the authors show how explicit modeling of predicate-argument structures can increase a transformer's capacity to detect paraphrases. In \cite{bai-etal-2022-semantic} it is shown how leveraging semantic structures during pre-training can improve a transformer's ability to understand complex dialogues. Such works motivate our desire to embed semantic graphs to alleviate the need for large-scale data annotation. Unlike previous works, we specifically aim to leverage semantics in a way that can be easily plugged into any text classification pipeline.

\subsection{Abstract Meaning Representation Graphs}  
Generally speaking, the goal of Abstract Meaning Representation (AMR) graphs is to capture ``who did what to whom" for a given text \cite{banarescu2013abstract}. More specifically, AMRs provide a semantic representation which \textit{abstracts away} from the text, producing a semantic graph with no 1-to-1 mapping to the original text, but rather a high-level encoding of the semantics necessary to preserve meaning. An example AMR graph can be found in Figure \ref{fig:example_parse}. Numerous studies in recent years have shown various use cases for AMRs in the domains of data augmentation \cite{amr_da}, dialogue modeling \cite{bonial2020dialogue}, sentiment classification \cite{li-etal-2022-cross-domain}, and semantic textual similarity \cite{cai-etal-2022-retrofitting}. 

Many AMR-based models involve complex architecture designs which enable the inclusion of AMRs in the text classification pipeline. This study differs in that we encode the AMR separately and simply concatenate the embedding to a text embedding. We aim to show that this simple model can significantly improve performance on low-resource health tasks with little engineering effort. 

In \cite{cai-etal-2022-retrofitting} they explore the retrofitting of AMR embeddings to text embeddings similar to our concatenate-then-predict format. However, their study does not explore end-to-end training of the AMR encoder and exclusively studies the impact of AMRs on multi-lingual tasks. We base our AMR encoder on the model introduced in this work. 

\begin{table}[t]
\renewcommand{\arraystretch}{1.1}
\begin{tabularx}{\linewidth}{X l l}

\toprule
\textbf{Text}  & \textbf{S-Complex} & \textbf{R-Complex}\\ 
\midrule
Is it safe to soak my earrings in rubbing alcohol (ethyl 70\%) everyday before putting them in \& putting them in right after?                    & 4.22 & 9.08  \\
\midrule 
Aspirin allergy - is it worth getting a bracelet? & 1.7                  & 6.70          \\
\midrule 
Authorities  have identified that the international chemical-warfare terrorist  ‘Samuel Whitcomb Hyde’ is behind the deadly China ‘coronavirus.’” & 5.48 & 14.37 \\
\midrule 
You shouldn’t open your windows                   & 1.5                 & 0.51          \\ 
\bottomrule

\end{tabularx}
\caption{Example syntactic complexity (S-Complex) and reading level (R-Complex) from four samples on various ends of the complexity spectrum. }
\label{complex_examples}
\end{table}

\subsection{Textual Complexity}

The notion of textual complexity is inherently subjective and requires further grounding to perform analysis. We choose to look at two statistical measures: 1) \textit{The Flesch Kincaid Reading Level} \cite{flesch1948new} is an estimate of what US grade level is required to read a given text. This metric is a function of the number of words per sentence and number of syllables per word. We find this metric to do a good job of separating short, simple texts from long, verbose texts and thus useful for this analysis. Most reading levels roughly fall within a range of 0-16 to represent  kindergarten (least complex) through university level (most complex) texts.  2) \textit{Syntactic Complexity: } Texts which are syntactically complex will have large distances between words which refer to one another in a sentence \cite{dep_distance_ref}. We obtain syntactic complexity by parsing each text into a dependency tree and computing the mean distance between parent and child nodes in the sentence. Most mean dependency scores fall in the range of 0-5, with higher numbers indicating samples with many co-referring tokens that are far from one-another in a given text. This metric is of interest as AMRs do a good job at capturing long-distance co-reference between nodes in multi-sentence texts. We compute both metrics using the TextDescriptives library \footnote{https://github.com/HLasse/TextDescriptives}. Examples of complexity statistics are displayed in Table \ref{complex_examples}.

\begin{table*}[!th]
\centering
\resizebox{\textwidth}{!}{%
\renewcommand{\arraystretch}{1.1}
\begin{tabular}{@{}llcc@{}}
\toprule
\multicolumn{1}{c}{\textbf{Tasks}} & \multicolumn{1}{c}{\textbf{Sample}} & \textbf{Label} & \textbf{Evaluation Metric} \\ \midrule
\textbf{Health Advice Detection} & Interventions to reduce self-harm in adolescents are needed. & Strong Advice & Macro F1 \\
\arrayrulecolor{gray}\midrule 
\textbf{COVID Rumor} & Weed kills coronavirus & False & Macro F1 \\
\arrayrulecolor{gray}\midrule 
\textbf{Medical Severity Detection} & I have pure pressures head aches and coughing persistent ? & Severe & Macro F1 \\
\arrayrulecolor{gray}\midrule 
\textbf{Medical Question Pairs} & \begin{tabular}[c]{@{}l@{}}Q1: Is hypno-therapy dangeorus?\\ Q2: Are effects of hypno-therapy permanent? I heard they can be dangerous. Is it true?\end{tabular} & Paraphrase & Macro F1 \\
\arrayrulecolor{gray}\midrule 
\textbf{Conditional Conflict Detection} & \begin{tabular}[c]{@{}l@{}}Advice 1: Limit dairy foods\\ Advice 2: If stomach upset occurs while you are taking this medication, you may take it with food or milk.\end{tabular} & Conflicting & Positive F1 \\
\arrayrulecolor{gray}\midrule 
\textbf{Temporal Conflict Detection} & \begin{tabular}[c]{@{}l@{}}Advice 1: Limit liquids before bed\\ Advice 2: Be sure to drink enough fluids to prevent dehydration unless your doctor directs you otherwise.\end{tabular} & Conflicting & Positive F1 \\
\arrayrulecolor{gray}\midrule 
\textbf{BIOSSES} & \begin{tabular}[c]{@{}l@{}}Sentence 1: The oncogenic activity of mutant Kras appears dependent on functional Craf\\ Sentence 2: Oncogenic KRAS mutations are common in cancer\end{tabular} & 2/4 & Spearman's Rank \\ \bottomrule
\end{tabular}%
}
\caption{Sample data from each of our 7 evaluation datasets. Health Advice Detection, COVID Rumor, and Medical Severity Detection are all single-text datasets. Medical Question Pairs, Conditional/Temporal Conflict Detection and BIOSESS are pairwise-inference tasks. The label for each sample and evaluation metric for each dataset are shown on the right. }
\label{DataExamples} 
\end{table*}

\section{Methods}

In this section, we describe the end-to-end process to obtain AMR representations of text. We then discuss the contrastive learning framework for embedding AMR graphs. Finally, we discuss how AMR embeddings are incorporated into a text classifier. 

\paragraph{Parsing \& Linearizing AMR graphs}
To obtain the AMR representations, we use the amrlib \footnote{https://github.com/bjascob/amrlib} library to parse each text into an AMR graph. Specifically, we use the `parse\_xfm\_bart\_base' model, which is a sequence-to-sequence parser based on the BART transformer \cite{lewis-etal-2020-bart}. In order to leverage the knowledge of pre-trained language models, we must convert the graph into a format which can be used by a transformer encoder. We do this by \textit{linearizing} the AMR graphs. This method is employed as opposed to operating on the graph directly, as it has been extensively shown to be the method of choice for many AMR-related tasks \cite{SPRING, mager-etal-2020-gpt}. To linearize a given graph, we employ a depth-first search-based linearization as done in \cite{cai-etal-2022-retrofitting}. An example AMR graph and it's corresponding linearization can be found in Figure \ref{fig:example_parse}. The resulting linearized graph can now be treated as a textual input in our classification pipeline. 

\paragraph{Embedding Linearized AMRs}
Once the AMRs are linearized, one can employ traditional sentence embedding strategies such as \textit{contrastive learning} to build meaningful AMR representation vectors. Contrastive learning works by constructing a dataset of triplets with the format (anchor, positive, negative), where the goal is for the model to push the embeddings of (anchor, positive) closer together and push (anchor, negative) further apart. This process produces meaningful semantic text embeddings that can be analyzed in high-dimensional space. Contrastive learning has been shown to be successful for many text embedding models \cite{reimers-2019-sentence-bert, gao2021simcse} as well as for multi-lingual AMR representations  \cite{cai-etal-2022-retrofitting}.  

\begin{figure*}[!h]
    \centering
    \includegraphics[width=\textwidth]{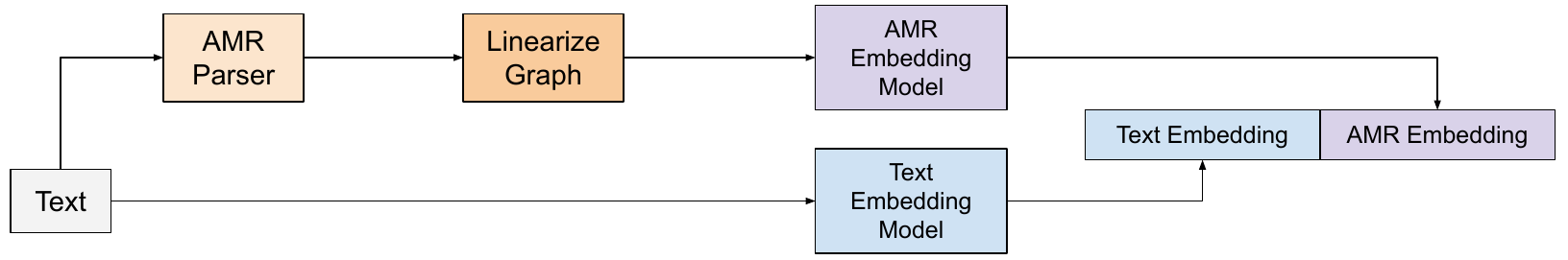}
    \caption{End-to-end pipeline displaying how each text gets $\rightarrow$ Parsed into an AMR graph $\rightarrow$ Linearized into a flattened string representation $\rightarrow$ Embedded by our AMR Embedding Module $\rightarrow$ AMR Embeddings are then concatenated with Text Embeddings and fed into a linear classifier. Note this example describes the process for a single-text input task. When the task is pairwise, additional steps in the pipeline occur to accommodate a second text. }
    \label{fig:pipeline} 
\end{figure*}

We construct a contrastive learning dataset using Natural Language Inference (NLI) data \cite{mnli, snli}. NLI is a pairwise inference task whose goal is to detect if Sentence B \textit{entails} or \textit{contradicts} Sentence A. This annotation serves as a surrogate contrastive triplet of the form (anchor, entailment, contradiction). The use of NLI for contrastive learning has become standard in the domain of sentence embeddings \cite{gao2021simcse}. Our general framework for AMR embedding is inspired by  \cite{cai-etal-2022-retrofitting}. To train our contrastive AMR embedding model, we first convert all 275,601 NLI training triplets into linearized AMR representations. We then fine-tune a PLM using the Sentence-Transformers library \footnote{https://www.sbert.net/} with contrastive learning using the Multiple Negatives Rankings loss \cite{multiple_rank_loss}. Our AMR-encoder uses the MiniLM architecture \cite{wang2020minilm} as it's backbone. We use the MiniLM model provided by the Sentence-Transformers library as it is a parameter efficient model (and thus more inclusive of future works with hardware restrictions) which has been pre-trained for sentence embedding tasks --- an initialization we found empirically useful in our experimentation \footnote{Huggingface Model: `sentence-transformers/all-MiniLM-L6-v2'}. We fine-tune our model using linearized AMRs for 1 epoch with a learning rate of $2e-5$ and early stopping using the STS-B development set \cite{cer-etal-2017-semeval}. Additionally, we use mean pooling and the native PLM tokenizer.

\paragraph{Merging Text \& AMR Embeddings at Inference Time}

In all experiments where we combine text and AMR information, our final representation is a simple concatenation of text and AMR embedding. Figure \ref{fig:pipeline} shows an end-to-end example of how texts become augmented by AMRs at inference time. We choose this approach as the purpose of this paper is to show that AMRs can be easily integrated into any NLP inference pipeline. We leave the application of more complicated AMR architectures [e.g. \cite{hyperbolic_amr, vgae_amr}] to future problem-specific applications.

\section{Evaluation Tasks}

In this section, we provide details about each of our 7 evaluation tasks. An example from each dataset can be found in Table \ref{DataExamples}. We additionally introduce the metric used to evaluate each dataset as well as provide intuition on why AMRs are useful to the modeling of the given task.

\subsection{Health Advice Detection (HAD)}

We explore Health Advice Detection (HAD) using the HAD dataset \cite{li-etal-2021-detecting}, which contains 5982 sentences from structured abstracts of biomedical research articles. The task is to detect if a sentence contains strong, weak, or no advice. We believe the HAD task will benefit from AMR's ability to model action polarities as well as common advice modifiers, such as conditionality and temporality. We evaluate performance on HAD by reporting the mean macro-F1 score over a stratified 5-fold cross-validation.  
\subsection{Health Conflict Detection (HCD)}

HCD was introduced in \cite{preclude}, where the task is to take two pieces of health advice from a public health resource and detect if and how they are conflicting. This task aims at protecting users with multiple pre-existing conditions from following advice that is true in the context of one of their diagnoses but dangerous or conflicting in the context of another. In \cite{gatto2023HCD}, they show that PLMs perform poorly on this task in part due to the complex multi-sentence nature of health advice data. 

In this study, we explore two HCD sub-tasks: Conditional Conflict Detection (\textbf{HCD-C}) and Temporal Conflict Detection (\textbf{HCD-T}). A \textit{conditional conflict} occurs when two pieces of health advice only disagree under a certain condition. A \textit{temporal conflict} occurs only when two pieces of advice only differ in terms of temporality. Examples of each conflict type can be found in Table \ref{DataExamples}. We believe AMRs are well suited to aid in the modeling of such texts as the AMR vocabulary explicitly models conditional and temporal relationships between texts. A good example of AMRs ability to model temporality can be found in Figure \ref{fig:example_parse}. Here we see the AMR identify that Epidemics happen with some ``frequency" and that ``every 100 years" is abstracted into a ``temporal quantity". 

Each HCD task is a binary classification task identifying if a pair of advice texts do or don't contain a conditional or temporal quantity respectively. We evaluate using HCD's 2825 train and 470 synthetic test samples.  Due to reasons outlined in \cite{gatto2023HCD}, we are unable to perform cross-validation on HCD as many pieces of advice are used in multiple pairs which could lead to data leakage. Thus, we report the mean F1 score of the positive class over 5 experimental trials. 

\subsection{COVID Rumor Detection}
The onset of the COVID-19 pandemic sparked a significant spike in online textual health misinformation \cite{kouzy2020coronavirus}. Detection of misinformation requires a complex understanding of world knowledge and semantic reasoning. Given AMRs use in argument modeling in existing literature \cite{opitz2021explainable}, we feel they may help identify the argument structure of erroneous claims. We evaluate AMRs impact on misinformation detection through the \textbf{COVID Rumors} dataset \cite{covid_rumor}. COVID Rumors contains a sub-task where the goal is to detect if a rumor is True, False, or Unverified. The dataset consists of 4129 claims from online news articles. An example False, or misinformative rumor can be found in Table \ref{DataExamples}. Samples were collected from various fact-checking websites where veracity was explicitly mentioned for each claim.  We evaluate performance on COVID Rumors by reporting the mean F1 score over a stratified 5-fold cross-validation. 

\subsection{Detecting Medical Question Duplicates }

The Medical Question Pairs (\textbf{MQP}) dataset \cite{MQP} sources questions from an online health community, HealthTap.com, to identify duplicate user queries. This is an important task in telemedical triage pipelines. MQP contains 3048 pairs of questions annotated for duplicate queries. For this task, doctors were presented with 1524 questions and asked to write similar and dissimilar pairs to produce a duplicate questions dataset. We argue that AMRs are valuable in the detection of questions with similar semantics and thus relevant to tasks like predicting question duplicates. We evaluate performance on MQP by reporting the mean macro F1-score over a stratified 5-fold cross-validation. 

\subsection{Classifying Severity of Telemedical Queries (Med-Severity)}

The Medical Severity Classification (\textbf{Med-Severity}) task \cite{severity_detection} contains 573 telemedical queries from online health communities such as HealthTap, HealthcareMagic, and iCliniq annotated for perceived severity. Similar to MQP, this is a crucial task in the telemedical triage pipeline. Med-Severity is a binary classification task where samples are annotated as either ``Severe" or ``Non-Severe". We evaluate performance on this dataset via the mean macro F1 over a 5-fold stratified cross-validation.

\subsection{Health-Specific Semantic Textual Similarity} 
We also evaluate how AMRs aid in health-specific semantic textual similarity (STS) through the lens of the  \textbf{BIOSSES} dataset \cite{BIOSSES}. BIOSSES is a biomedical STS dataset containing 100 sentence pairs from biomedical research literature. Annotation is done on a scale of [0-4], with 0 meaning sentences have no similarty and 4 meaning they are semantically equivalent. AMRs have been shown to aid related STS tasks in the literature \cite{cai-etal-2022-retrofitting}. As mentioned in our discussion of MQP, AMRs, being they are a semantic structure, should be able to aid in the modeling of semantic relatedness. 

Unlike our other experiments, BIOSSES is evaluated in a completley unsupervised setting. We first map the labels to be between [0-1] and then compute the cosine similarity between each pair of vectors. This is the standard approach for STS evaluation in the literature \cite{gao2021simcse}. We report the Spearman's Rank correlation between all the cosine similarities and ground truth annotations as our BIOSSES evaluation metric.

\begin{table*}[t]
    \begin{center}
    \centering
    \small
    
    \begin{tabular}{lcccccccc}
    \toprule
       Model                 & HCD-C & HCD-T & MQP & COVID Rumor & HAD & Med-Severity & BIOSSES  & Avg. \\
    \midrule
    \midrule
        \multicolumn{9}{c}{\it{Static Embeddings}}\\
    \midrule
        SBERT               & 0.20 $\pm.13$          & 0.69 $\pm.003 $           & $\mathbf{0.69 \pm.01}$   & $0.64 \pm.04$   & 0.68 $\pm.01  $          & 0.86 $\pm.03$          & 0.81 & 0.64 \\
        AMR-Only            & $\mathbf{0.41 \pm.02}$ & 0.72 $\pm.004 $           & 0.59 $\pm.03$            & 0.62 $\pm.04 $           & 0.73 $\pm.01 $           & 0.85 $\pm.03$          & 0.76 & 0.65 \\
        SBERT+AMR           & 0.34 $\pm.006$         & $\mathbf{0.74 \pm.003}$    & 0.67 $\pm.03$            & $\mathbf{0.66 \pm.04}$   & $\mathbf{0.76 \pm.01}$   & $\mathbf{0.88 \pm.01}$ & $\mathbf{0.83}$ & $\mathbf{0.67}$ \\
     
    \midrule
        \multicolumn{9}{c}{\it{Dynamic Embeddings}}\\
    \midrule
        SBERT                &  $0.79 \pm.01$          & 0.83 $\pm.02$            & $\mathbf{0.77 \pm.04} $          & $\mathbf{0.71 \pm.05}$ & $\mathbf{0.89 \pm.01}$ & 0.88 $\pm.03$          & -  & 0.81 \\
        AMR-Only             & 0.67 $\pm.01$           & 0.78 $\pm.02$            & 0.67 $\pm.05  $         & 0.64 $\pm.03 $         & 0.84 $\pm.01  $        & 0.87 $\pm.03$          & -  & 0.74 \\
        SBERT+AMR            & $\mathbf{0.82 \pm.01}$  & $\mathbf{0.85 \pm.02}$   & $\mathbf{0.77 \pm.06}$  & $\mathbf{0.71 \pm.04}$ & $\mathbf{0.89 \pm.01}$ & $\mathbf{0.89 \pm.02}$ & -  & 0.82 \\

    \bottomrule
    \end{tabular}
    \end{center}

    \caption{
        Each column describes performance of Text-Only, AMR-Only, and Text+AMR models for each evaluation task. In the static evaluation setting, we find AMRs improve performance on 6/7 evaluation tasks. In the dynamic evaluation setting, AMRs only improve performance on 3/6 tasks. Note that BIOSSES is not applicable in the dynamic setting as STS is evaluated completely unsupervised. 
    }
    
    \vspace{-5pt}
    \label{table:results}
\end{table*}

\section{Evaluation Setting} 

We perform each experiment with both \textit{static} and \textit{dynamic} embeddings. We define a static embedding experiment as one where learning occurs \textit{only} in the classification head (i.e. text and AMR encoders are frozen). A dynamic embedding experiment is where the text and AMR encoders can be updated during training, i.e., they are  \textit{not frozen}. Evaluating in both contexts is important for low-resource learning as small datasets can overfit in end-to-end fine-tuning settings. So it is useful to analyze performance in the static setting where the pipeline is more regularized.

In each experiment we use the MiniLM-based SBERT  \cite{wang2020minilm} architecture  provided by the Sentence-Transformers library \cite{reimers-2019-sentence-bert}  as it is one of their top performing, most popular sentence encoders which has been pre-trained on over 1-billion sentence pairs \footnote{Huggingface model string: sentence-transformers/all-MiniLM-L6-v2}. 
 We choose this PLM as it is extremely accessible in terms of both availability and number of parameters. We use the same encoder backbone for both text and AMR models, as well as static and dynamic experiments to maintain fair and consistent evaluation across different evaluation settings. 

All static embedding experiments are performed by taking a fixed embedding and feeding it into a linear classification head. We train the linear layers for 5 epochs with a learning rate of 0.001 using the AdamW optimizer with weight decay = 0.01 to reduce overfitting. All classification experiments use the standard Cross Entropy loss with a balanced class weight. 

Dynamic embedding experiments are similarly performed by passing embeddings into a linear classification head. However, since we allow gradient updates to occur in the encoders, we fine-tune for 5 epochs with a learning rate of $5e-5$. All other training parameters are the same as the static experiments.

\section{Results}

\subsection{Static Embeddings}

We find that all of the advice datasets (i.e. HCD-C, HCD-T, HAD) find static AMR embeddings useful for text classification. This validates our claim that advice texts have explicit semantic structures which can be exploited by AMR graphs in text classification pipelines. A particularly interesting result is that an AMR-only model with no access to the text shows higher performance than text-only models on both the HCD-C and HCD-T tasks. This is likely due to AMRs capacity to model explicit conditional and temporal relationships in the data. Patterns amongst labels and AMR relations such as \textit{:condition} and \textit{:time} may have provided a useful signal towards the detection of a conditional or temporal conflict.  This result motivates future work which aims to connect AMRs to HCD in a more task-specific manner. 

We find that AMRs do not help the MQP dataset in the static setting. This result is surprising as duplicate questions should have similar AMRs. However, it may be too challenging for the AMRs to be compared in the linear classification head of the static classifier, causing poor results on this task. On average, AMR-infused models aided in predictions of the COVID Rumor dataset. However, each run had relatively high variability as shown in Table \ref{table:results}, thus limited conclusions can be drawn from such experiments. On the Med-Severity dataset, AMRs were shown to provide a performance boost of 2 F1-pts. Given the extremely low-resource nature of Med-Severity, this task benefits mainly from the additional modality, as severity detection has less overt connections with semantics and greater ties to world/medical knowledge. 

Finally, we find that improved performance in the BIOSSES task follows other results in the literature where AMR improves results on STS \cite{cai-etal-2022-retrofitting}. That is, simple concatenation of text and semantic embeddings can improve performance on unsupervised cosine-similarity tasks such as BIOSSES, even without any domain-specific pre-training.

\subsection{Dynamic Embeddings}

Our experimental results show that the only tasks with a conclusive boost in performance from AMR-infused PLMs are the HCD tasks. Again, AMRs are particularly well-suited for such tasks given the overlap between the HCD label space and the AMR relation vocabulary. On other inference tasks such as MQP, COVID Rumor, and HAD, we do not find AMR-infused models to be effective in the dynamic setting. We are again surprised by the result that AMRs to not aid in the detection of medical question duplicates. Future works should explore if domain-specific pre-training or deep learning architectures which are built to \textit{compare} AMR graphs for text classification may make them useful for this task. Both COVID Rumor and HAD saw the same performance with and without AMR embeddings. Since AMRs were extremely helpful in the static setting for HAD, it may be the case that AMRs are only useful when advice modeling is in an either lower-resource or more regularized state (e.g. static embedding evaluation). We can confirm this suspicion by re-running the dynamic experiments in an artificially lower-resource setting by randomly sampling 500 examples from HAD and re-running the experiment. Our results in Table \ref{extremely_low} confirm that, in the extremely low resource setting, AMRs to in fact help the dynamic HAD experiment. They do not, however improve performance on extremely low-resource variants of MQP or COVID Rumor. In general, low-performance on COVID-Rumor is likely due to the nature of veracity prediction as it is extremely dependent on world knowledge. Future works may wish to explore incorporating AMRs into knowledge-infused PLM architectures.

\begin{table}[!th]

\centering
\resizebox{\columnwidth}{!}{%
\begin{tabular}{@{}llll@{}}
\toprule
Model     & MQP         & COVID Rumor & HAD         \\ \midrule
SBERT     & $\mathbf{0.68 \pm.06}$ & $0.60 \pm.05$ & $0.75 \pm.04$ \\
AMR-Only  & $0.54 \pm.04$ & $0.55 \pm.08$ & $0.75 \pm.02$ \\
SBERT+AMR & $0.66 \pm.06$ & $\mathbf{0.61 \pm.04}$ & $\mathbf{0.80 \pm.01}$ \\ \bottomrule
\end{tabular}%
}

\caption{Results of the dynamic experiments on a random subset of 500 samples from each of our three largest datasets. This experiment aims to investigate if AMRs are more useful in an extremely low-resource setting. Our results find that HAD shows significant improvement from AMR embeddings when only presented with 500 samples. The other two tasks, however, still show no conclusive evidence that they benefit from semantic graph embedding. }
\label{extremely_low}

\end{table}

\section{Why Do AMRs Improve Performance on Low-Resource Health Datasets?}
\begin{figure*}[t]
    \centering
    \includegraphics[width = \textwidth]{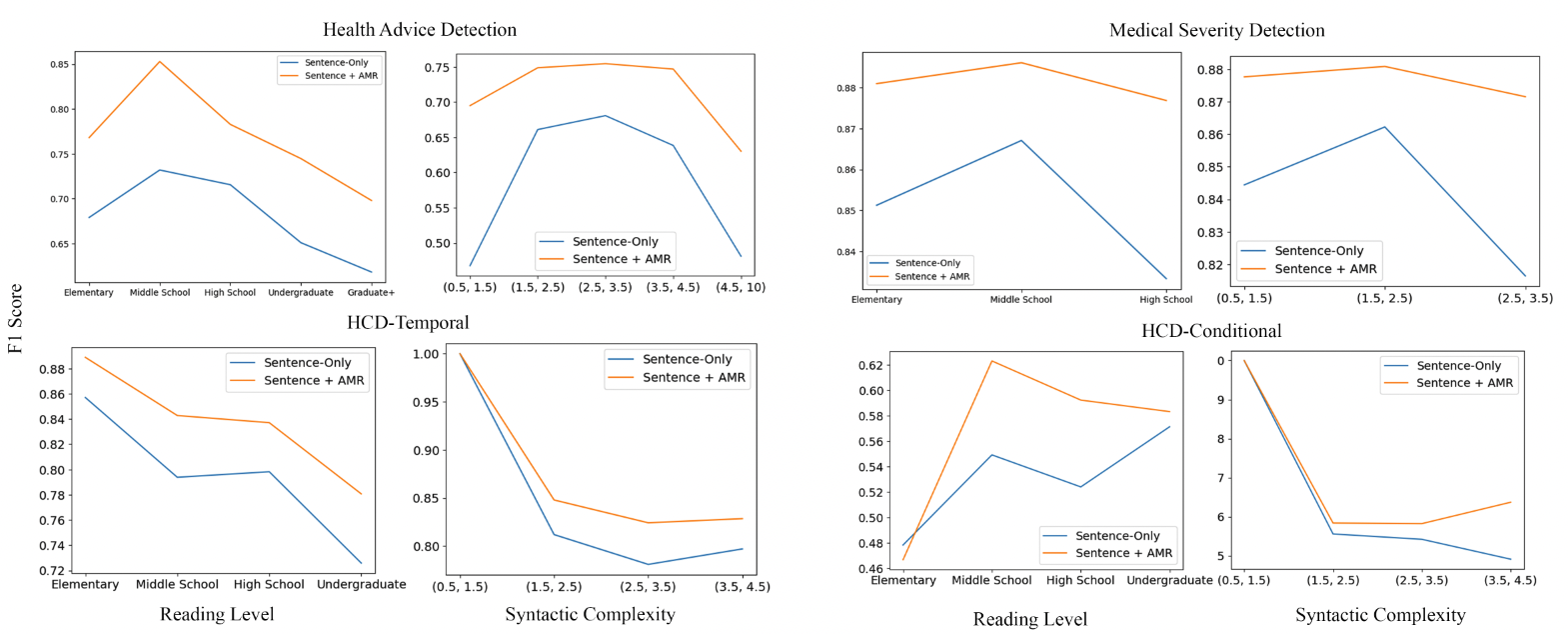}
    \caption{Analyzing how the F1 score relates to reading level and syntactic complexity on four datasets using the static evaluation setup. The x-axis denotes the textual complexity bins (higher = more complex). Specifically, the x-axis of reading level experiments are binned by US grade levels. The x-axis of syntactic complexity experiments are binned by mean dependency distance.  Our results show that AMR-based models perform better and show less predictive variance on the most complex health texts. }
    \label{fig:err_analysis}
\end{figure*}
In this section, we provide an error analysis of the four static experiments where we saw the greatest improvement from the inclusion of AMR embeddings --- HAD, HCD-C, HCD-T, and Med-Severity. We are specifically interested in investigating our claim that AMRs improve performance on complex health texts, which can be both multi-sentence and filled with challenging vocabulary.

\paragraph{Why Analyze Complexity? } 
We choose to perform error analysis through the lens of complexity as AMRs are by design abstracting away from individual words and into a smaller set of high-level concepts, producing a simpler representation of textual semantics. Our intuition is that if AMRs are doing a good job at representing a sentence using only relations between semantic concepts, it should make the modeling of difficult texts more efficient. However, it is important to mention that these two metrics are not a perfect measure of \textit{sample difficulty}. For example, the Flesch Kincaid Reading Level considers sentences with many high-syllable words as being more complex. However, one could argue that the high-syllable adjective ``significant" is easier to understand in context than the noun ``bank", as one could be referring to a river bank, a bank shot, or the Bank of America. However, our empirical analysis found this to be a reasonably good way to separate extremely easy texts from the extremely difficult ones. In other words, we believe there is a functional difference in sample complexity when we look at two samples that are very far away on our defined complexity spectrum and thus they serve as a useful measure of text complexity. In summary, this analysis doesn't necessarily mean we are improving performance on more \textit{difficult} samples (i.e. difficult for transformers to understand), but it does show how performance changes through the lens of different linguistic attributes often associated with what humans understand to be complex texts.

\paragraph{Experimental Setup: } In order to evaluate how Reading Level and Syntactic Complexity affect performance, we conduct the following experiment: First, we take the predictions from our SBERT Sentence-Only static experiment and compare it to the Sentence + AMR variant. For HAD and Medical Severity Detection, we look at all predictions from all 5 test sets from each fold in our 5-fold cross-validation. For the HCD tasks, we look at the mean prediction from each of our 5 experimental trials. For each text, we compute each measure of complexity. We then split the predictions into bins based on complexity. For each sample bin, we recompute the macro F1 score for samples in that bin and plot them in Figure \ref{fig:err_analysis}. 

\paragraph{The Relationship Between F1 Score and Syntactic Complexity: } Figure \ref{fig:err_analysis} shows that AMR-infused models perform better across-the-board on texts with the highest syntactic complexity. We specifically find an interesting pattern of performance divergence between text-only and AMR-infused models as the complexity of texts increases. For example, on the HCD-T task, both models perform equally well on easy texts, but start to diverge as texts get more and more complex. A similar pattern is found on the Med-Severity and HCD-C tasks, where there is a clear divergence in F1 between the last two complexity bins. On the HAD task, while we don't see the same divergence pattern, we do notice that the AMR-infused model has much less variability with respect to changes in syntactic complexity when compared to text-only models. We can thus conclude that AMRs make PLMs more robust to syntactically complex health texts in low-resource settings. 

\paragraph{The Relationship Between F1 Score and Reading Level: } Our results show that, for HCD-C, HCD-T, and HAD, Reading Levels for both Sentence-Only and Sentence+AMR models follow similar trends. That is, changes in complexity seem to affect performance for Sentence-Only and Sentence+AMR models similarly. However, AMR-infused PLMs perform better on texts with higher reading levels in general. On the Med-Severity dataset, we find a significant drop in F1 between middle school and high school level texts when compared to the AMR-infused model which remains relatively stable. From this experiment, we can only conclude that AMRs do in fact perform better on texts with higher-reading levels. However, we do not observe the same divergence trend found in the syntactic complexity experiment, as AMR-infused models lose performance at a similar rate to text-only models as the reading level increases.

\section{Broader Impacts}

\paragraph{Seamless Integration of AMR Embeddings For Any Downstream Health NLP Task: } This study improves performance on low-resource health tasks while maintaining an easy-to-implement classification technique. The success found using a simple concatenate-then-predict infrastructure means this work can be easily extended to any future low-resource Health NLP tasks with little engineering effort. Additionally, we release our model using the popular Huggingface / Sentence-Transformer libraries which allow users to produce AMR embeddings with very few lines of code. 

\paragraph{Reducing The Need for Large-Scale Annotation of Health Web Texts: } Our work shows that with semantic modeling, we may be able to reduce the need for large-scale annotation of training samples for Health NLP tasks. This result extends the problem space solvable by encoder-based PLMs, specifically augmenting the scope of PLMs for web and social media texts. 

\paragraph{Complex Texts Become More Accessible to PLMs: } Our work shows that AMR embeddings make complex health texts easier for PLMs to process. This is important as social media datasets from sources like Twitter, Facebook, and Reddit often produce multi-sentence samples. Thus, explicit modeling of multi-sentence texts is crucial for the advancement of social media-based text classification pipelines. 

\section{Ethical Impact}

This research involved no human subjects as experiments were all run on publicly available data. The selection of datasets chosen aimed to represent a diverse and relevant set of Health NLP tasks. However, it may be the case that the datasets chosen contain sample distributions that under-represent certain population groups. For example, the MQP dataset contains 1524 unique questions, which is small compared to the list of possible medical ailments one may inquire about. Future works which leverage AMR embeddings for medical duplicate detection should ensure performance stays consistent across different types of medical conditions not reflected in the MQP dataset. Similarly, samples in the conflict detection dataset are from official public health sources which target the general population and may not reflect health conflicts common among smaller underrepresented subgroups. Any deployment of an AMR-based solution to low-resource health tasks should be aware of such data biases. 
\section{Limitations \& Future Work}
\paragraph{Limitations: } Any AMR-based model which depends on silver-labeled AMR parsings of texts is inherently limited by the performance of the AMR parser. In this paper, the parser we employed achieves an 82.3 SMATCH \cite{smatch} score on the AMR LDC 2020 dataset. This parser shows high performance, and we may see AMR embedding results improve as AMR parsing algorithms become more accurate. However, given that there are potential errors in the AMR parsing process, we may find that difficult samples produce bad semantic embeddings, which may introduce errors into the system. This problem could potentially become more prevalent as dataset complexity increases. 

Our evaluation framework has additional limitations as some tasks (e.g., HCD-T, HCD-C) are not constructed to use a hold-out evaluation set for training without encountering data leakage. Similarly, other tasks, such as Med-Severity, are extremely low-resource ( $< 1000 $ samples) and are thus too small to permit a hold-out validation set. These challenges motivate using a standardized set of training parameters, but this may not produce the optimally performing model for each experiment. Similar discussions on low-resource validation sets can be found in the literature \cite{kann2019towards}.

\paragraph{Future Works: } Future work on AMR embeddings for health tasks may explore domain-specific pre-training strategies. In this work, we only leverage NLI data for AMR embeddings, but improvements may be found via in-domain pre-training for any of our evaluation datasets. For example, unsupervised pre-training of our AMR encoder on COVID-19 tweets may have improved performance in the static embedding experiments for COVID Rumor. 

Additionally, future works may explore more complicated social media sources such as Reddit, where posts can often be multiple paragraphs long. Gold-labeled AMRs have annotations for many samples with multiple sentences, making them potentially valuable for modeling paragraphs. Evidence for AMRs use in tasks with longer, more complicated texts are evident in dialogue modeling \cite{bonial2020dialogue} and thus may apply to longer Reddit posts. 

\section{Conclusion}

The challenges associated with coordinating large-scale human annotation of health texts, such as time constraints, access to resources, and access to domain experts, are unlikely to go away soon. We must take explicit steps towards crafting low-resource solutions for texts in the health space as they are crucial to implementing many safety-critical public resources. In this work, we introduce AMR embeddings in the context of low-resource health tasks and show how they can help increase performance without large-scale datasets. We additionally show that AMRs are helpful in modeling complex health texts found on the web and online health communities, which are often complicated multi-sentence text with varying degrees of nuance.

\bibliography{aaai23}

\end{document}